# Using Artificial Intelligence to Analyze Fashion Trends

**Mengyun (David) Shi**     **Van Dyk Lewis**

Cornell University,  Hearst Magazines

ms2979@cornell.edu

**Abstract**

    Analyzing fashion trends is essential in the fashion industry. Current fashion forecasting firms, such as WGSN, utilize the visual information from around the world to analyze and predict fashion trends. However, analyzing fashion trends is time-consuming and extremely labor intensive, requiring individual employees' manual editing and classification. To improve the efficiency of data analysis of such image-based information and lower the cost of analyzing fashion images, this study proposes a data-driven quantitative abstracting approach using an artificial intelligence (A.I.) algorithm. Specifically, an A.I. model was trained on fashion images from a large-scale dataset under different scenarios, for example in online stores and street snapshots. This model was used to detect garments and classify clothing attributes such as textures, garment style, and details for runway photos and videos. It was found that the A.I. model can generate rich attribute descriptions of detected regions and accurately bind the garments in the images. Adoption of A.I. algorithm demonstrated promising results and the potential to classify garment types and details automatically, which can make the process of trend forecasting more cost-effective and faster.

**Keywords:** artificial intelligence, fashion, trend analysis

**Note: this paper provides an introduction of how A.I. can be adopted for fashion trend analysis. If you are working or studying in the fashion industry. this is the right paper for you to get started.**

**However, if you are an advanced computer/research scientist, we suggest you read the fashionpedia project that we collaborate with Google AI directly:** https://fashionpedia.github.io/home/index.html

1. Introduction

Fashion attribute detection has received increasing attention in computer vision communities in recent years (Bossard, 2017). The percentage of fashion related papers using the proven scientific method (Pont-Tuset, n.d.) has increased by 20% from 2010 to 2017 at the top computer vision conferences, Computer Vision and Pattern Recognition (CVPR) and International Conference on Computer Vision (ICCV), which indicated a steadily traction in this area during the past few years. There have been a large body of research involving clothing modeling, attributes recognition, image parsing, image retrieval, and styling recommendations. As an essential task for consumers in fashion e-commerce, image retrieval gained particular interests recently. The majority of the research work focused on retrieving daily life images and online shopping ones (Hidayati, Hua, Cheng, & Sun, 2014; Vittayakorn, Yamaguchi, Berg, & Berg, 2015), as it is, only two articles targeted on runway images; neither of it detecting fashion attributes from professionals' perspective who work in the fashion industry. Fashion attributes detection for the "fashion insiders" requires a more detailed and in-depth analysis of fashion images than that for consumers.

Fashion attribute detection from images is critical for professionals in fashion industry, especially for trends forecasting. For example, fashion forecasting firms, such as WGSN, detect fashion attribute information from images all around the world (fashion shows, visual merchandising, blogs, and streets) (Banks, 2013). They examine the information through experiences, observations, media scans, interviews, and exposure to new places. Such information analyzing process is known as abstracting: it recognizes similarities or differences across all the garments and collections (Fiore & Kimle, 1997). Fashion forecasters abstract information across design collections and time intervals to identify changes in fashion trends.

Designers, product developers and buyers abstract information across garment groups or collections to develop a cohesive and visually appealing lines. Sales and marketing executives abstract information across product lines of each season to recognize selling points. Fashion journalist and bloggers abstract information across runway photos to deliver symbolic core concepts that can be translated into editorial features (Bloomsbury.com, n.d.). Curators from museums and libraries abstract across historical fashion photos and collections to archive them in digital format. Fashion scholars abstract through both historical and modern fashion photography for their own research purpose.

    Analyzing image-based fashion attributes usually is time consuming and labor intensive, even for professionals. For instance, WGSN employs around 150-200 forecasters to analyze the trends (Banks, 2013) for each season regularly. Furthermore, the trend is changing much faster than before. Timely and reliable trend forecasting becomes much more important than before to meet consumers' needs. Manual labeling and labor intensive classification of visual information limit the speed and reliability in interpreting fashion dynamics to all over the world. In addition, manual labeling based on each individual's experience is quite subjective due to human errors or opinions in certain fashion attributes. A possible breakthrough of taking advantage of artificial intelligence that has been widely adopted in other fields but not fashion field could improve fashion forecasters' analysis and classification of visual information through providing systematic, and computerized methods.

    Considering the financial cost in manual labeling and the accuracy in classifications based upon human subjective judgment, this explorative study proposes a data-driven quantitative abstracting approach using an artificial intelligence (A.I.) algorithm. The algorithm could potentially improve the efficiency and lower the cost of analyzing fashion images through

the following: .firstly, an A.I. model was trained to be familiar with images from a large-scale dataset under different scenarios, such as online stores and street snapshots; secondly, the model could detect garments and classify clothing attributes such as fabric textures, garment style, and design details from runway photos and videos; thirdly, the model could summarize fashion trends from the attributes it developed. The proposed hypothesis are:

> **H1:** The proposed A.I. model will generate accurate fashion attribute descriptions from detected regions in the runway images.
>
> **H2:** The proposed A.I. model will precisely summarize fashion trends from the given runway images.
>
> **H3:** The proposed A.I. model will identify fashion attributes from runway photos and videos.

## 2. Literature review

### 2.1 Detecting clothing categories and attributes from fashion images

Earlier works relied on some classical computer vision approaches to sort apparel into categories and describe fashion attributes in images (Bossard et al., 2012; Bourdev, Maji, & Malik, 2011; H. Chen, Gallagher, & Girod, 2012; Di, Wah, Bhardwaj, Piramuthu, & Sundaresan, 2013). Further research work was dedicated to a computer vision technique called "image segmentation" for classifying different apparel categories via probabilistic methods (Simo-Serra, Fidler, Moreno-Noguer, & Urtasun, 2014; Yamaguchi, Kiapour, & Berg, 2013; Yamaguchi, Kiapour, Ortiz, & Berg, 2012). In order to improve the accuracy in recognizing garments, some of the research mentioned above contained preprocessed images that accounted for human posture and body parts (Figure 1(a)).

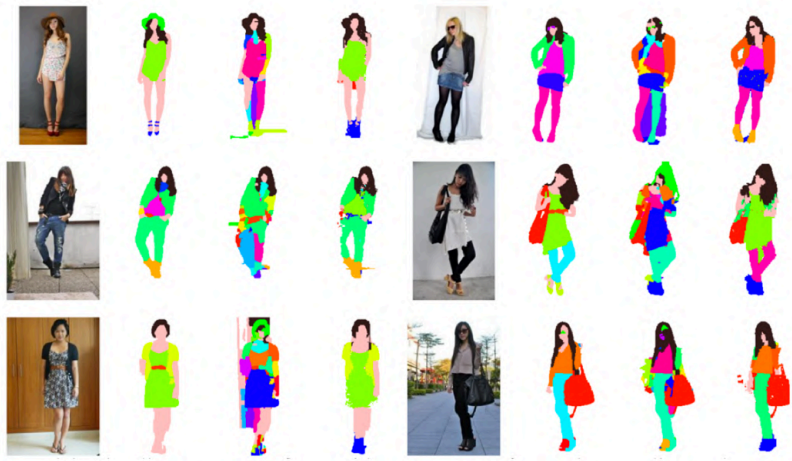

(a) Recognizing garments by estimating human posture and quantifying body parts

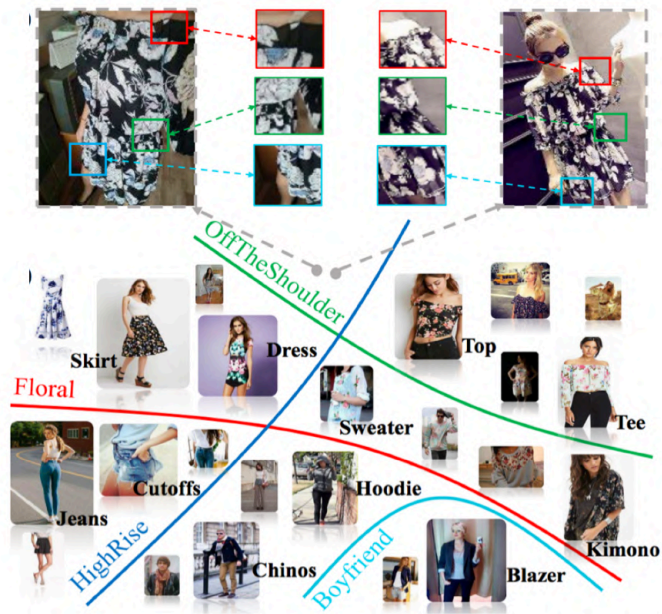

(b) The A.I. algorithm called R-CNN was used to detect garment parts and generate attribute description

Figure 1. The previous methods used to detect clothing categories and attributes from fashion images (Yamaguchi, Hadi Kiapour, & Berg, 2013; Liu, Luo, Qiu, Wang, & Tang, 2016).

In addition, some computer vision researchers (Hadi Kiapour, Han, Lazebnik, Berg, & Berg, 2015; Vittayakorn et al., 2015) focused on retrieving images that have high similarity in fashion attributes. Other researchers began to use more advanced A.I. approaches to tackle this problem because it showed a significant boost to the accuracy in detecting fashion attributes. For example, many researchers (Chen et al., 2015; Dong, Gong, & Zhu, 2017; Huang, Feris, Chen, & Yan, 2015) started to utilize an A.I. algorithm called 'R-CNN' to detect garment parts or generate attribute description in fashion images (Figure 1(b)). The method proposed in this study is mostly based on an augmented version of A.I. algorithm called 'Faster R-CNN' (Ren, He, Girshick, & Sun, 2015). Compared to all previous methods, Faster R-CNN algorithm has been proven to be able to provide more accurate detection result (Ren et al., 2015) with rich attribute descriptions of garments in fashion images.

## 2.2 Analyzing fashion trends based on fashion attributes

Early work on trend analysis (Hidayati et al., 2014) broke down catwalk images from NYC fashion shows to locate style trends in high-end fashion. The recent advance in A.I. algorithm enabled more work in this area. Several recent approaches (He & McAuley, 2016; Matzen, Bala, & Snavely, 2017; Simo-Serra, Fidler, Moreno-Noguer, & Urtasun, 2015) utilized the A.I. algorithm to extract clothing attributes from images and create a visual embedding of clothing style cluster to investigate the fashion trends in clothing around the globe. However, it is not quite clear whether the trends discovered in these studies are accurate or not. Therefore this study further evaluated these approaches by comparing the fashion trends summarized from both the A.I. algorithm and the world's leading fashion trends magazine Vogue, ELLE, and Harper's BAZAAR ("The Very Best Fashion Magazines, Ranked," n.d.; "Top 10 Fashion Magazines," n.d.; "Top 13 Fashion Magazines In The World," n.d.).

## 2.3 Fashion images retrieval from different domains

There have been a number of works tackling the issue of cross-domain fashion images retrieval. The most popular topic in this area is the retrieval of similar fashion images from different domains such as street and online shopping (Cheng, Wu, Liu, & Hua, 2017; Gu et al.,

2017; Ji, Wang, Zhang, & Yang, 2017), which is extremely useful for e-commerce fashion websites. Most of the works in retrieving images is based on the current advance of A.I. approach (Kuo & Hsu, 2017; Wang, Sun, Zhang, Zhou, & Jiang, 2016), where it is necessary to teach the A.I. algorithm to recognize attributes shared by similar fashion items from both street and online shopping domain (Dong et al., 2017; Huang et al., 2015).

## 3. Method

### 3.1 A.I. augmented fashion trend analysis conceptual framework

A fashion trend analysis conceptual framework using the proposed A.I. method was proposed (Figure 2(a)). The framework was divided into two stages: 1) Detecting fashion attributes using A.I.; 2) Analyzing fashion trends based on fashion attributes detected in Stage 1. The evaluation methods were also provided at each stage to test the effectiveness of the proposed A.I. system. Furthermore, the proposed A.I. system was adopted to conduct a more challenging task - tracking fashion models in runway videos.

(Note: The results produced by the proposed A.I. system were further evaluated by comparing it with the fashion trends summarized by the fashion editors from the world's leading fashion magazines Vogue, ELLE, and Harper's BAZAAR. See supplementary materials for more details of comparison results.)

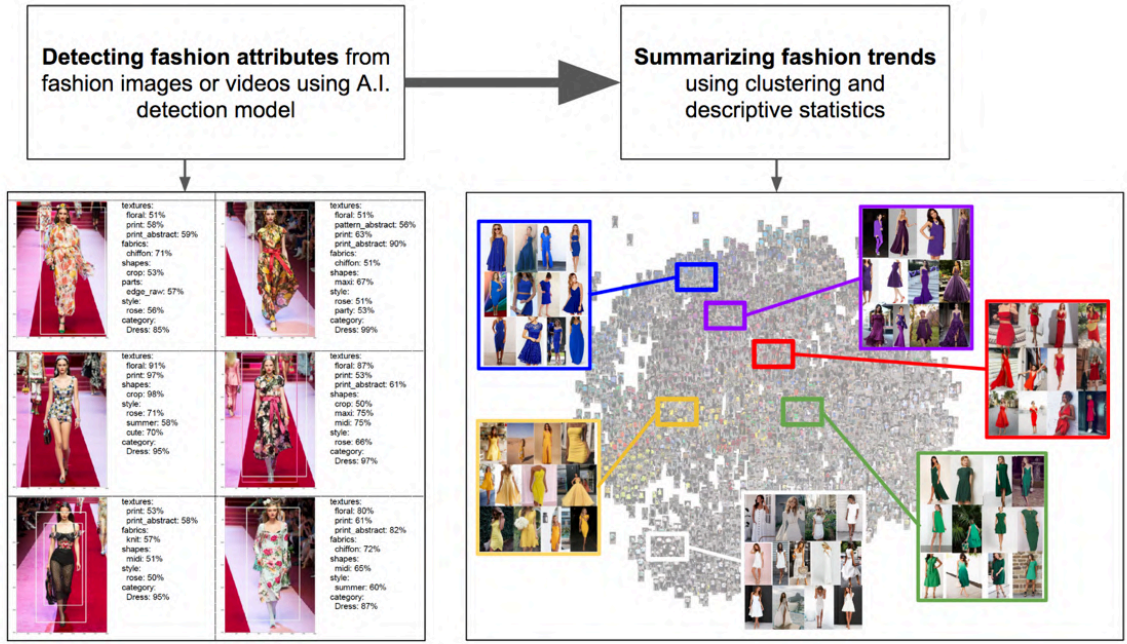

(a) Overview of the A.I. augmented fashion trends analysis methods.

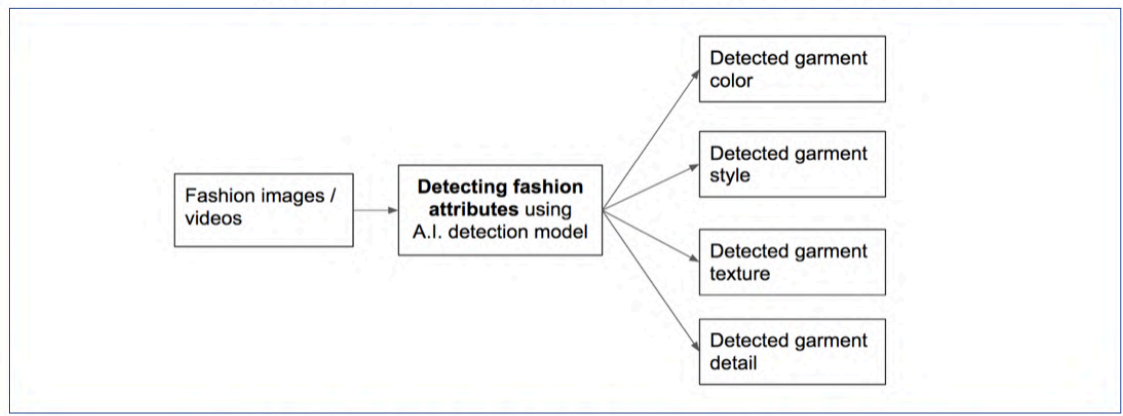

(b) Stage 1 – the conceptual framework for detecting fashion attribute.

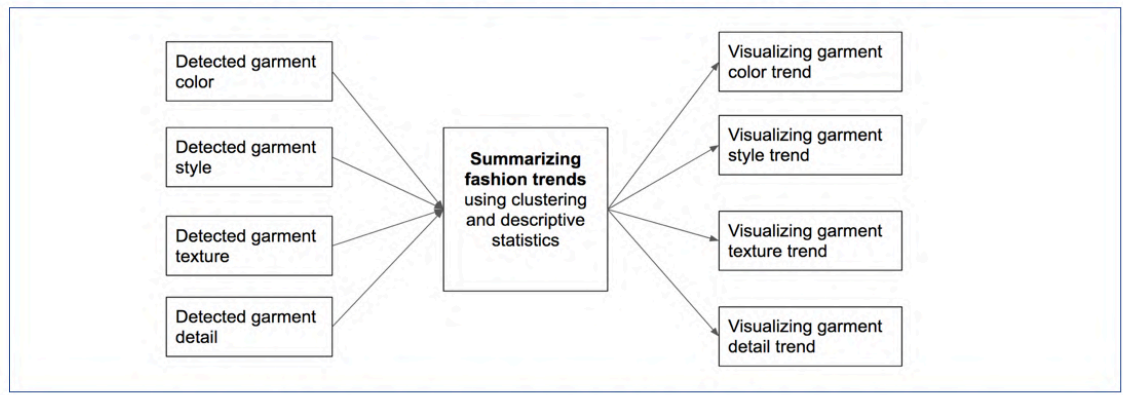

(c) Stage 2 - the conceptual framework for analyzing fashion trends.

Figure 2. The A.I. augmented fashion trends analysis methods.

### 3.1.1 Detecting fashion attributes using AI

In order to use the proposed A.I. method for analysis of fashion images, the A.I. algorithm had to go through a process called "training". During the training process, the A.I. algorithm was taught to learn the concept of fashion attributes from an image dataset annotated with certain fashion attributes (Figure 2(b)). After that, the A.I. algorithm could start to predict clothing attributes in new images. For instance, the A.I. algorithm learned the concept of 'maxi dress' from the annotated dataset and then predicted the 'maxi dress' in millions of unlabeled images. In this study, a large-scale fashion dataset called 'DeepFashion' was adopted since it contains over 280,000 images gathered from online shopping websites and street photos and relatively comprehensive annotations of garment details (Liu, Luo, Qiu, Wang, & Tang, 2016). Meanwhile, a powerful A.I. algorithm called 'Faster R-CNN' was also utilized to leverage the DeepFashion dataset (Ren et al., 2015). In order to test the effectiveness of the proposed A.I. system, the A.I. algorithm was used to analyze 16 looks from the Dolce & Gabbana Spring 2018 ready-to-wear collection. The testing result is interpreted in the result section.

Furthermore, our A.I. model run on 1844 runway images across 46 collections of the Spring/Summer 2018 presented in New York, London, Milan and Paris. The results of our A.I. model can be found in the supplementary material section.

### 3.1.2 Analyzing fashion trends

The goal of this stage was to identify fashion trends in a certain season by summarizing the fashion attributes detected from the last stage. In order to do so, both the popular machine learning algorithm called 'clustering' and the descriptive statistics methods were adopted. (Figure 2(c)). Clustering is a statistical technique used in fields such as data mining, pattern recognition, information retrieval and image analysis. It can allow a set of objects with similar properties to be automatically grouped. In this study, clustering was used to evaluate the fashion attributes across the images, and then to regroup the similar fashion attributes to the same groups (clusters). By analyzing these clusters, the popular fashion attributes could be discovered. Furthermore, the proposed method could also be used to analyze fashion trends beyond these attributes in a number of ways, enabling some new visual insights as follows:

1. To identify and analyze visually correlated fashion attributes in terms of four perspectives: garment color, garment style, garment texture and garment detail (Figure 2 (c)). So, we might address questions such as what is the most popular color and garment style among these images?

2. To visualize the trendy fashion elements in each season or year by analyzing a much larger corpus of images (spanning years) to a certain time period. For instance, we might find leggings are very popular a few years ago. In order to visualize these fashion elements, images need to contain information related.

3. To specify which cities are more correlated to each cluster. For example, we might find Canada Goose jacket is extremely popular in New York City last year, compared to Los Angeles. The prerequisite of building such clusters is that images have to contain both time period and location information. The clustering algorithm could distinguish images and visualize trends according to this two type of information.

To test the effectiveness of the proposed framework at this stage, women's products from ZARA.com was adopted since ZARA is recognized as the world's largest apparel retailer. Analyzing fashion attributes from the existing products in ZARA could give us a brief overview of current fashion trends around the world. The A.I. algorithm developed at Stage 1 was used to analyze all ZARA women images (around 10,000 products). Then, the detected fashion attributes were regrouped and summarized using clustering and descriptive statistics to abstract fashion trends.

### *3.1.3 Comparing the fashion trends summarized from the A.I. algorithm with the predictions from the world's leading fashion magazines*

The predictions of season Spring/Summer 2018 presented in New York, London, Milan and Paris made by fashion editors from Vogue, ELLE, and Harper's BAZAAR are used to compare with our A.I. model results. See the supplementary material section for more detailed results and comparison.

### 3.2 Analyzing fashion show videos

This study also further evaluated the proposed A.I. system by conducting a more challenging task: analyzing runway videos. Unlike static fashion images, runway videos consist of more complicated contents and scenes, which make it more difficult to be analyzed. Furthermore, tracking moving fashion models in fashion shows is also very challenging since it requires the proposed A.I. system to track fashion models in videos frame by frame. In order to prevent the proposed A.I. system from losing track of moving fashion models in runway videos, a technique called "image segmentation" was adopted in this study. Image segmentation is a technique used in the computer vision field typically to locate objects and boundaries (lines, curves, etc.) in images automatically. Figure 3 shows one example of segmentation technique. It was used to locate the outlines of the person, the sheep, and the dog in the image automatically (Lin et al., 2014).

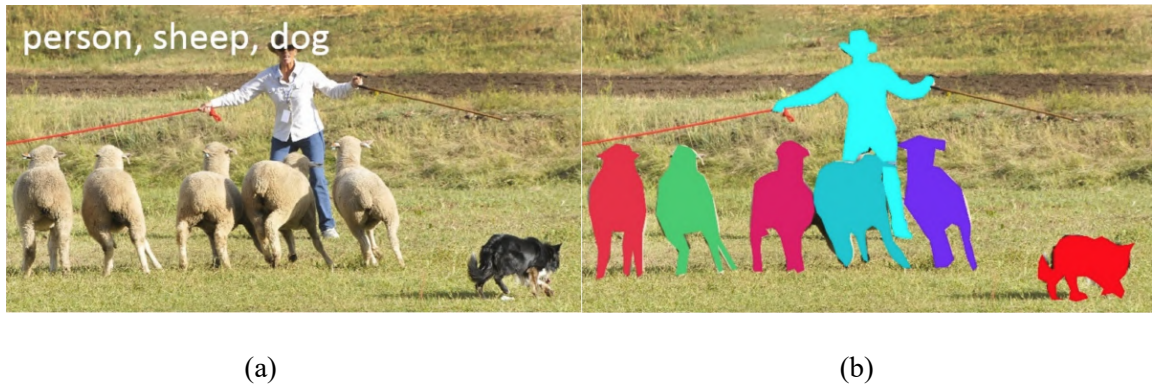

(a)                                                      (b)

Figure 3. (a) the original image ; (b) "image segmentation" technique is used to locate the outlines of the person, sheep and dog in the image (Lin et al., 2014).

The video of the Burberry Prorsum Fall 2012 Menswear Collection was used to evaluate the effectiveness of the proposed A.I. system in tracing moving fashion models.

## 4. Results

**4.1 The result of fashion attributes detection**

Figure 4 presents some of the promising predictions of the A.I. model. The model generated rich attribute descriptions of detected regions and accurately bounded the garments in the images. For instance, the model correctly identified 8 images belonging to the dress category with high confidence (higher than 85%). The styles included rose, summer, cute and party, in agreement with the "the sweet life" lifestyle the brand constantly celebrates. It correctly detected 6 floral patterns (one with lace), 5 patterns as abstract; the shapes of the dresses were either maxi or midi. The model predicted one dress as both maxi and midi with same confidence (75%) (Figure 4-1), this was mainly due to the length of the dress was in between knee and ankle (Figure 4-1). The model also detected two dresses were the bodycon (Figure 4-2 and Figure 4-6). This result is very promising because the classic hourglass shape and corseted tailoring are one of the designers' signature looks.

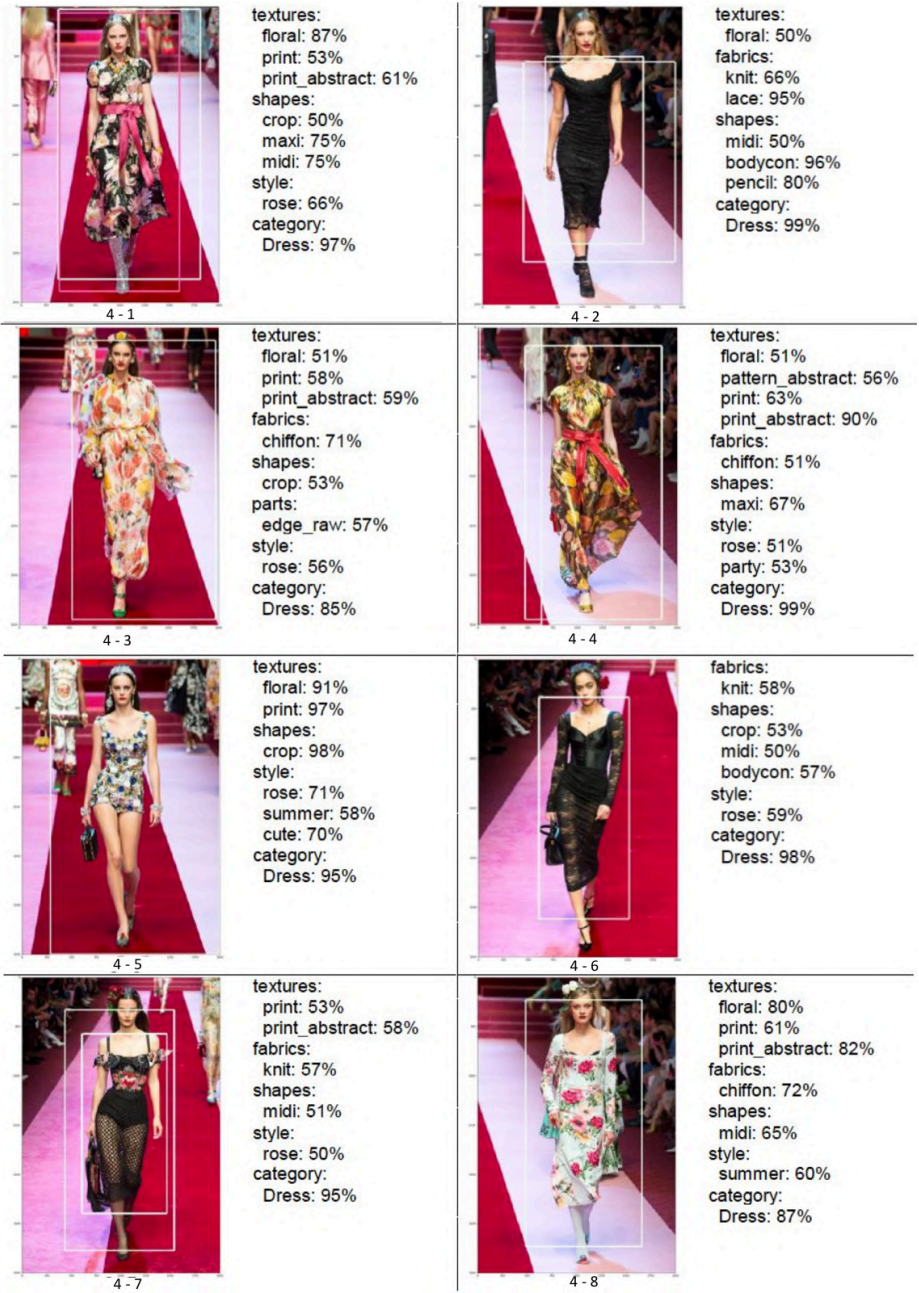

Figure 4. Example images using the proposed A.I. model. Each number after an attribute represents how confident the model's predictions are.

However, the A.I. model was not perfect. Figure 5 features some of the detected images with predictions that were not fully correct.

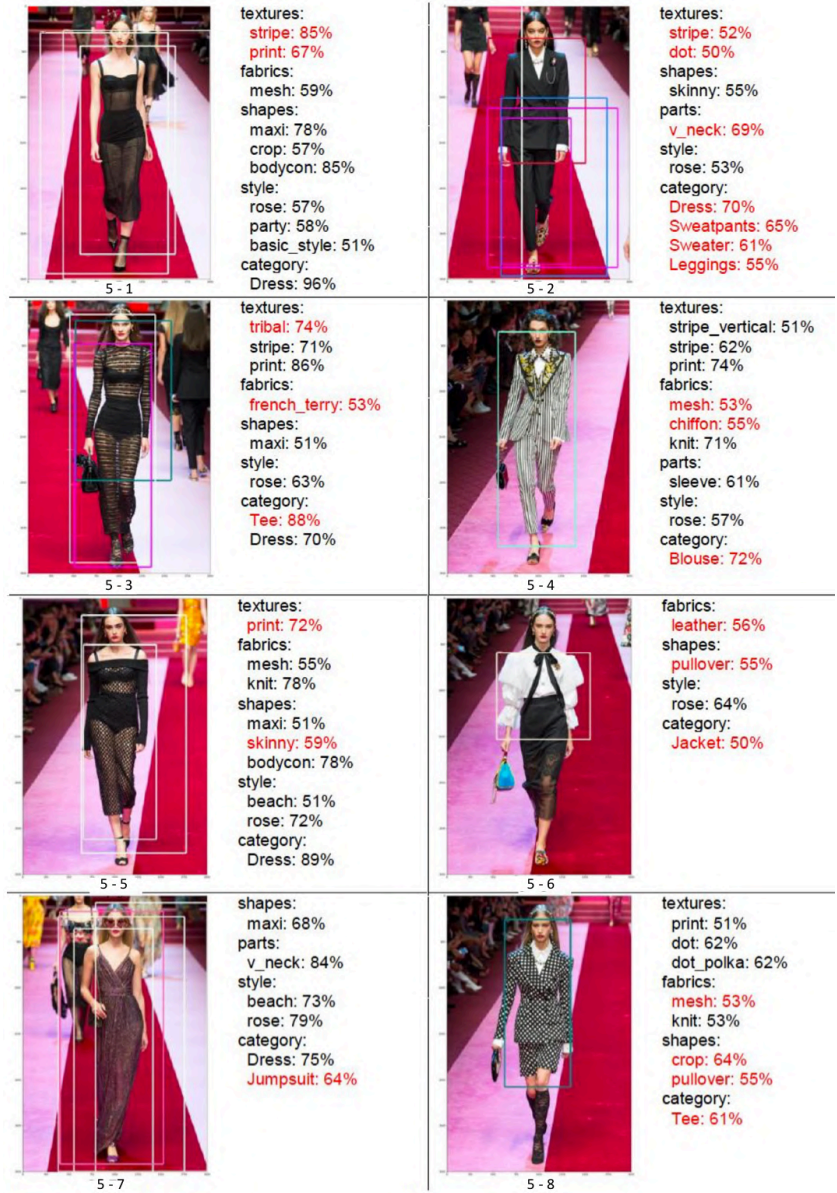

Figure 5. Example images using the proposed A.I. model. Each number after an attribute represents how confident the model's predictions are. The attributes highlighted in red represent incorrect predictions.

The model behaved poorly in detecting blazers, trousers or skirts. For example, the pantsuit in Figure 5-2 was incorrectly detected as dress, sweater, leggings, and sweatpants. The see-through mesh dress at Figure 5-3 was detected as a tee. The A.I. model was also confused from similar shapes. One example was, the front opening of the blazer at Figure 5-2 was wrongly classified as V-neck due to the similar V-shapes. Another example was in Figure 5-7, a maxi glitter dress was identified as a jumpsuit.

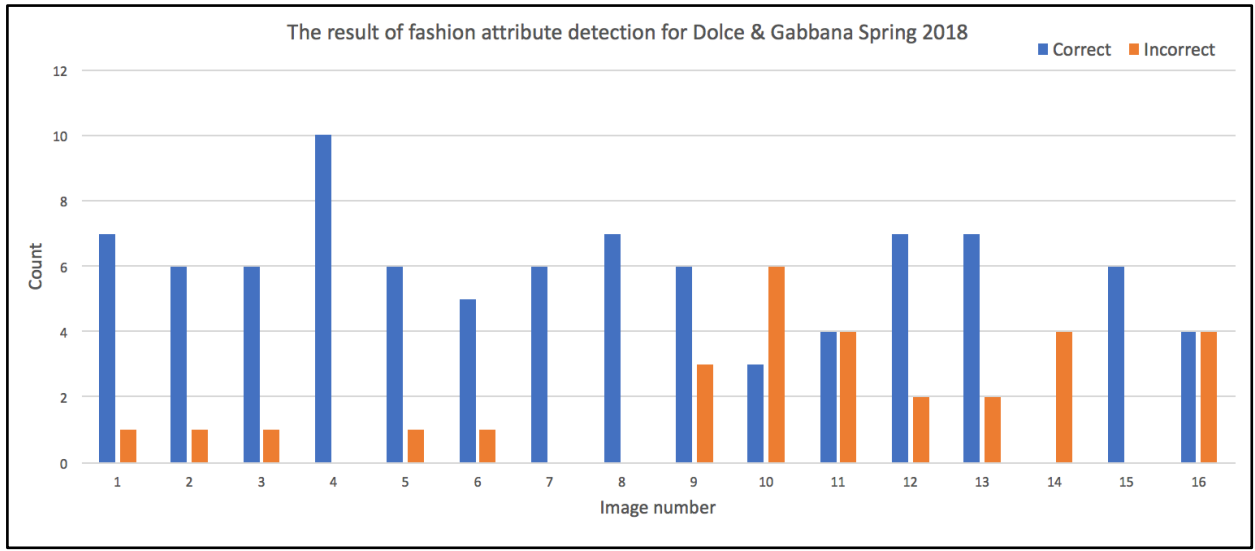

Figure 6. The results of fashion attribute detection produced by the proposed A.I. model.

The statistical analysis on the fashion attributes built from the 16 looks in the Dolce & Gabbana (Figure 6) shows that 75% prediction made by A.I. model is correct and 25% prediction is incorrect. It partially supports the hypothesis that the proposed A.I. model can generate accurate fashion attribute descriptions from detected regions in the images.

**4.2 The result of trend analysis**

*4.2.1 Garment style & detail trends*

Since ZARA images contain time period information but not location information, this study could only investigate fashion elements which represent the hottest trends in a certain season (Summer 2018 when this study was conducted). Figure 7 presents some of the results in analyzing garment style & detail trends from ZARA. Tops, T-shirt, and dress have the highest frequency in 'Garment style' plot (Figure 7(a)). Short sleeves appeared as the second most frequent fashion attributes in 'Garment detail' plot (Figure 7(b)). These results can be interpreted as a seasonal trend since all the products analyzed in this study belong to Summer 2018 collection. It also seems to be reasonable because ZARA might intend to sell more garments that consumers could wear for that summer season.

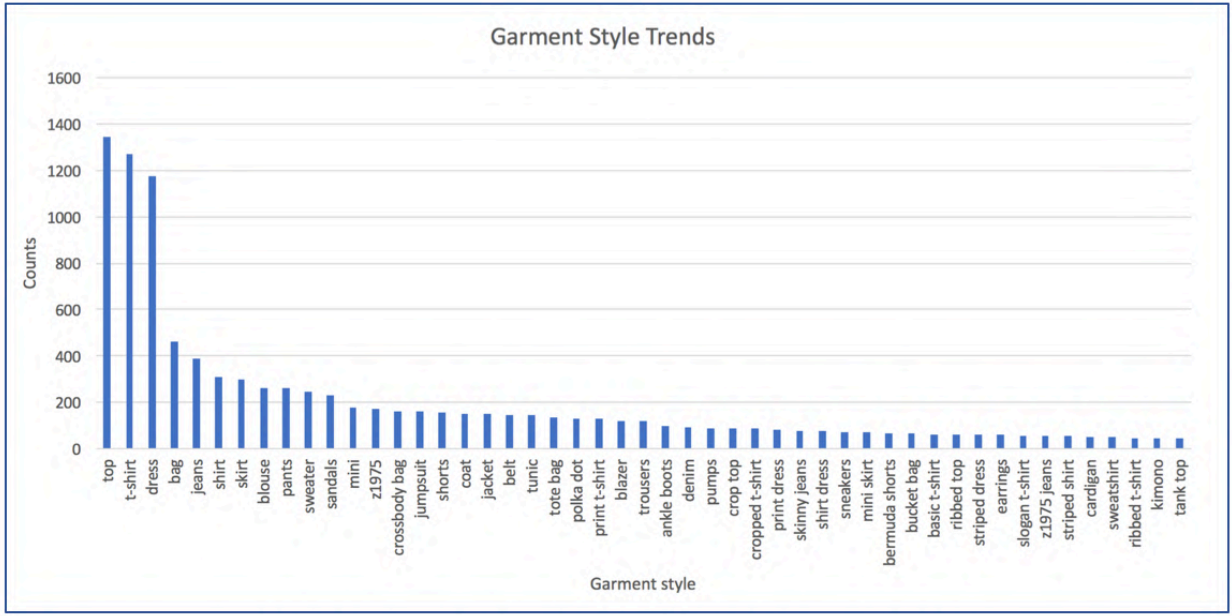

(a) ZARA garment style trends produced by the proposed A.I. model

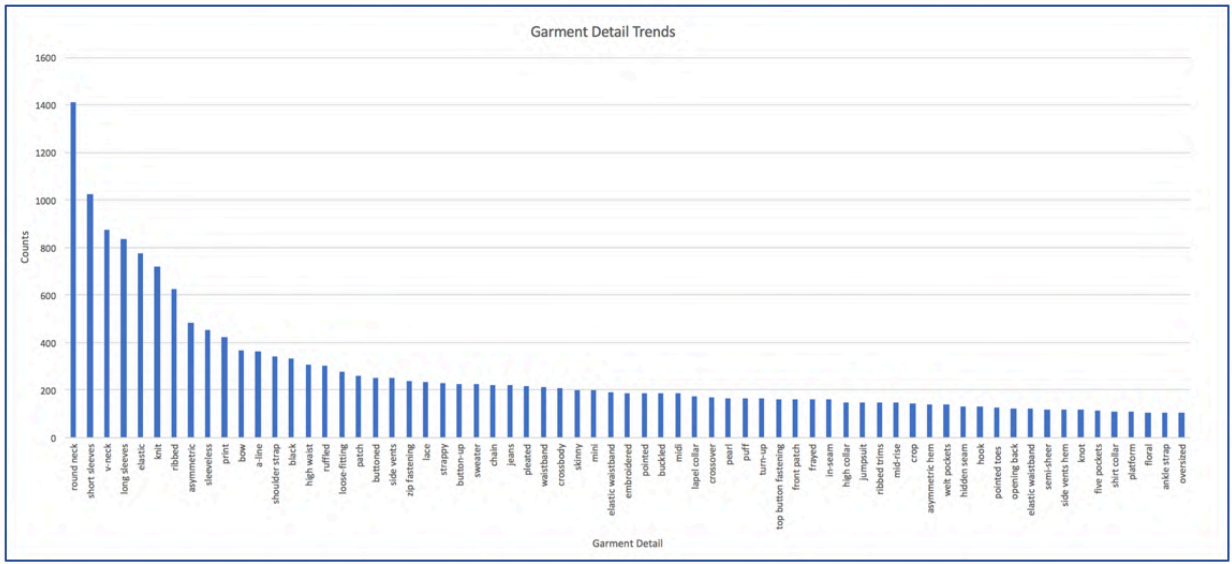

(b) ZARA garment detail trends produced by the proposed A.I. model

Figure 7. The result of the garment style & detail trends

Furthermore, Summer 2018 fashion trends from Vogue ("Summer 2018 fashion trends," 2018) was used as a secondary source of information to evaluate the accuracy of the trend results summarized by the proposed A.I. model. The result showed that both Vogue and the A.I. model listed oversized, floral, shirtdress and polka dot as the popular fashion trends for Summer 2018. The attributes such as sequined, baby doll, lightweight, check & plaid and patterned were also suggested by Vogue as the summer trends. However, these attributes were not found in the trend report produced by the A.I. model. One possible explanation is that the proposed A.I. model might not be accurate enough to detect every fashion attribute from ZARA products. Another possible reason might be that ZARA did not include these attributes in their mood board while designing their summer season product line, which made it impossible for the A.I. model to detect these attributes.

*4.2.2 Color trends*

Figure 8(a) presents some of the results in color trends analysis from ZARA. White and black occupied the top 2 positions at the 'Color trends' plot. However, it is safe for not interpreting it as Summer 2018 trends because consumers usually wear these two colors for all seasons, not only for Summer 2018. Trendy colors should vary over seasons.

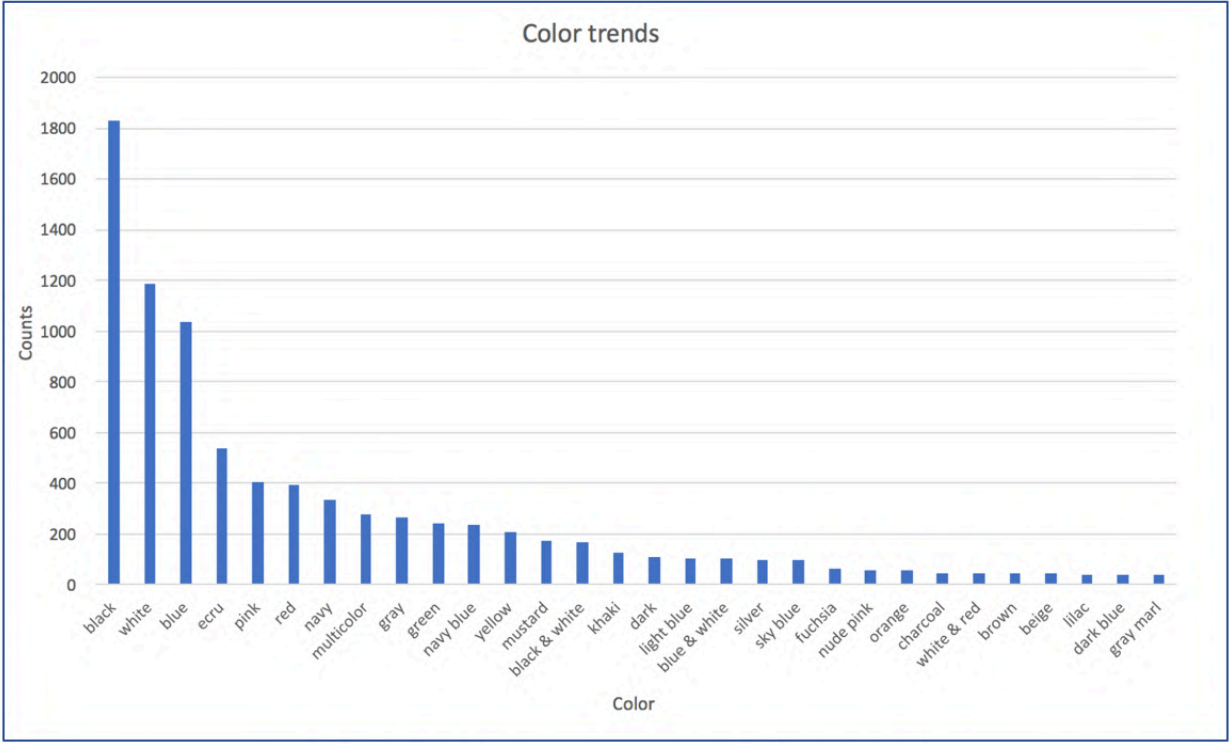

(a) ZARA color trends produced by the proposed A.I. model

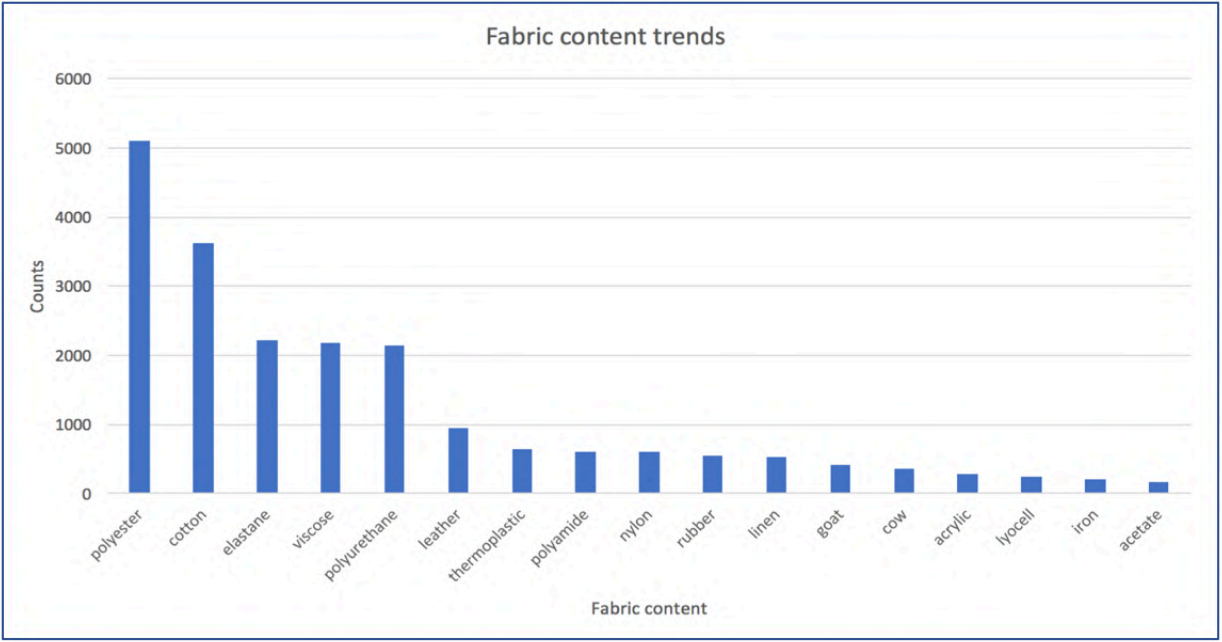

(b) ZARA fabric content trends produced by the proposed A.I. model

Figure 8. The result of the color and fabric content trends

Similar to garment style & detail trends analysis, Summer 2018 color trends from Vogue ("The ten color trends for Spring Summer 2018," 2018) was used as a secondary source of information to evaluate the accuracy of the trend results analyzed by the proposed A.I. model. Unlike the results of garment style & detail, the color trends analysis exhibited some interesting but less obvious connections. For instance, Vogue suggested the following color as the leading trends for Summer 2018: sky blue, tomato red, light green and military green. Findings from the proposed A.I. model indicated that blue, red and green were discovered as the most popular colors. However, it was not quite clear which shade of blue or red or green represented the color trends for Summer 2018. Further investigation on the color attributes from ZARA products was conducted. It was found that ZARA did include sky blue color in some portion of their products (Figure 8(a)). However, it is hard to say whether ZARA considers sky blue as the trendy color for their Summer 2018 product lines, because only 98 products among 10,000 products were found containing sky blue attributes. For the red color, it was confirmed that ZARA did not produce any products that were tomato red, . but some products that were white & red (47 products) for the summer season. For the green color, ZARA did not design any products associated with light green and military green. The results of the investigation showed that the inaccurate results in color trends analysis were mainly caused by the inconsistent information between Vogue and ZARA, not the proposed A.I. model itself.

The results of fabric content analysis from ZARA is also shown in Figure 8(b). The study did not further evaluate the accuracy of this result since Vogue did not fail to provide any trends report related to fabric content for Summer 2018.

## 4.3 The result of runway video analysis

Figure 9 and Figure 10 present some results using segmentation techniques to track the fashion models in the Burberry Prorsum Fall 2012 Menswear fashion show.

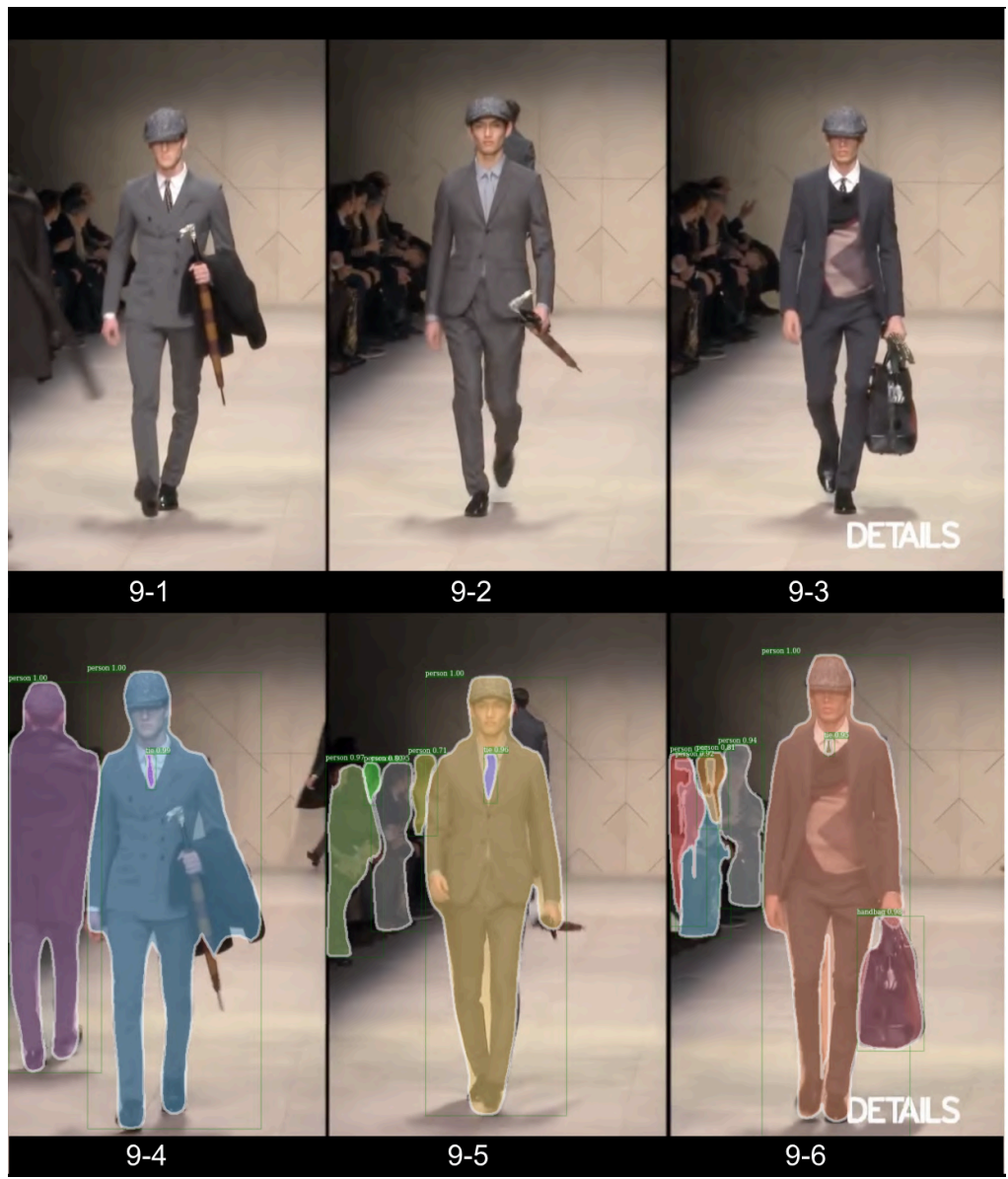

Figure 9. Tracking fashion models using "image segmentation" technique - 1. The upper row is the original video frame. The lower row is the result using segmentation technique. The video adapted from Burberry Fall 2012 Menswear Runway Recap. (n.d.). Retrieved September 24, 2018 from https://www.youtube.com/watch?v=w2YCUbf86zA.

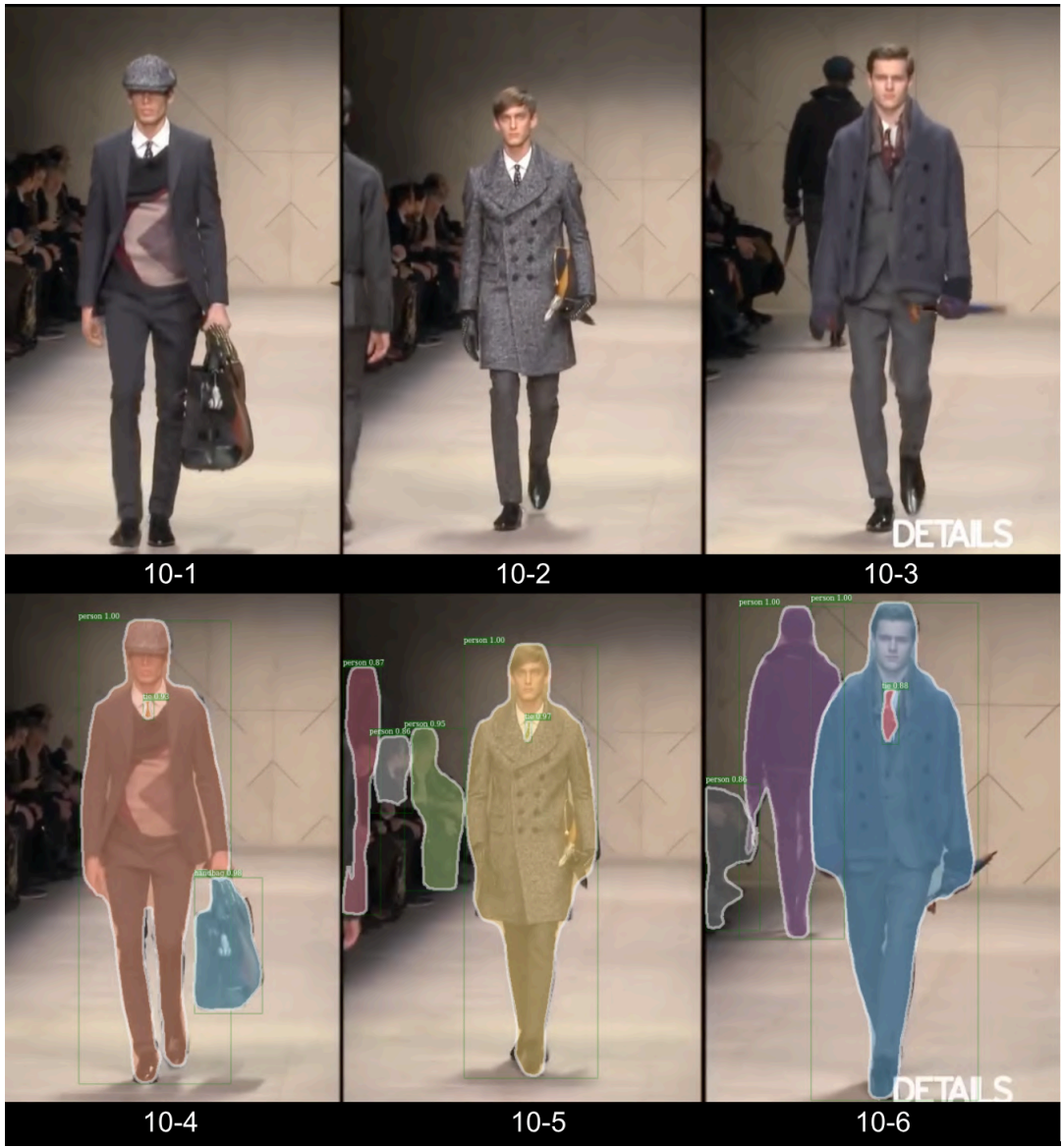

Figure 10. Tracking fashion models using "image segmentation" technique - 2. The upper row is the original video frame. The lower row is the result using segmentation technique. The video adapted from Burberry Fall 2012 Menswear Runway Recap. (n.d.). Retrieved September 24, 2018 from https://www.youtube.com/watch?v=w2YCUbf86zA.

      The A.I. model could accurately track and bind the male fashion models and their garments in the images even if the fashion models were moving quickly. Meanwhile, it could also detect the accessories such as ties, handbags, and umbrellas carried by the fashion models. However, The A.I. model had trouble in recognizing other garment items such as blazers, coats, and pants. Furthermore, the tracking quality of the fashion models became a little bit worse when

the fashion models walked in a row at the end of the show (Figure 11). One explanation for this possible problem is due to the overlap of these fashion models confused the A.I. model to distinctly recognize each fashion model and their garments.

Figure 11. Tracking fashion models using "image segmentation" technique - 3. The upper row is the original video frame. The lower row is the result using segmentation technique. The video adapted from Burberry Fall 2012 Menswear Runway Recap. (n.d.). Retrieved September 24, 2018 from https://www.youtube.com/watch?v=w2YCUbf86zA.

## 5. Discussion

Hypothesis 1 was proved by the fact that the proposed A.I. model can generate accurate fashion attribute descriptions from detected regions in the images with the average precision around 75%. Hypothesis 2 was partially supported due to only two garment style trends from Vogue's reports match the results of ZARA garment style trends produced by the A.I. model. It is also observed that only two garment detail trends and four color trends proposed by Vogue can be found from the corresponding ZARA detail and color trends analysis produced by the A.I. model. However, these results warrant further research into evaluating the trend analysis produced by the A.I. model in a more comprehensive way. Hypothesis 3 was somewhat established based on the A.I. model could tract and bind the male model with moving images, detect some garments and accessories successfully. The main contribution of the research compared with previous literature includes the following:

***More accurate A.I. model & datasets.*** No doubt, artificial intelligence is a powerful approach in analyzing visual information. However, both our results and previous work (Chen et al., 2015; Dong, Gong, & Zhu, 2017; Huang, Feris, Chen, & Yan, 2015) show that the A.I. approach can be improved from two areas: 1) A better A.I. algorithm is needed for detecting fashion attributes with higher accuracy. The current advance from research frontiers shows that the improvement of A.I. algorithm is still very promising. For instance, the AI model developed by Google achieves 93.2% accuracy, surpassing the human-level score of 91.2% ("Open Sourcing BERT," n.d.); 2) The quality of the image datasets needs to be improved as well. In our study, a large-scale fashion dataset called 'DeepFashion' was adopted to teach the A.I. model the concept of fashion attributes (Liu et al., 2016). However, around 60% of the fashion attributes annotated in 'DeepFashion' dataset contain errors. This is one of the important reasons causing the proposed A.I. algorithm fails to predict clothing attributes correctly. Thus, if our proposed A.I. model were to be used by fashion professionals in analyzing fashion images, we would need new datasets with correct annotations annotated by fashion experts, . to help the A.I. model understand the concept of fashion attributes better, as proposed in Fashionpedia project ("Fashionpedia," n.d.).

*Segmentation techniques.* The qualitative results show that the quality of the segmentations generated by our A.I. model (Figure 9-11) is better than the previous work (Yamaguchi, Kiapour, & Berg, 2013), as shown in Figure 1(a). It is mainly because this study utilized a more advanced A.I. model proposed by the researchers from the Facebook AI Research (He, Gkioxari, Dollár, & Girshick, 2017). However, the proposed A.I. model can only detect the basic apparel categories such as tie, bags, and umbrella. Future work should explore how to build an A.I. model to understand more comprehensive apparel categories ("Fashionpedia," n.d.).

*Image types.* Previous work on trend analysis (Matzen et al., 2017; Simo-Serra et al., 2015) utilized the A.I. algorithm to extract clothing attributes from daily life and street images to investigate the fashion trends in clothing around the globe. Our study broke down catwalk and product images from fashion companies to figure out style trends. However, both directions could be useful for fashion professionals and companies. We would propose to unifyand explore both directions in the future.

*Comprehensive comparison.* The comparison between Vogue and ZARA might be somewhat unreasonable. Vogue trend reports target on high-end luxury brands and products; ZARA products are more in line with low-end fast fashion. In the future study, comparing the trends analysis between Vogue and luxury brands such as Louis Vuitton and Gucci might reveal more meaningful insights.

# 6. Conclusion, Limitations and Future Work

Based on results from this exploratory study, the proposed A.I. model is believed to have ability to detect fashion attributes from images within some degree of accuracy (around 75%). The proposed A.I. model is able to discover garment details that cannot be easily seen by human eyes with the ability to run 24/7 without tiredness; it can be easily scaled up to detect fashion attributes, perform on a large number of images – identify not only similarities but also differences among varied garments, and connect runway to "real-ways" or vice versa. Hence, "fashion insiders" such as fashion forecasters and designers do not have to spend hours in analyzing fashion images, producing trend reports and preparing for product development manually.

It also indicates that there is a need for further investigation of trend analysis produced by the proposed A.I. model. While the developed A.I. model is not yet perfect, adoption of an A.I. model has demonstrated the promising potential to provide fashion designers, trend forecasters, fashion media editors and buyers with several benefits:

1) It can certainly lessen the financial and environmental burden for professional fashion companies and organizations;

2) It can discover garment details that is not easily seen by human eyes;

3) It can help trend forecasting firms and fashion companies work in a more efficient way.

The current fashion trends are changing faster and more complicated than ever. Many fashion bloggers are creating their own fashion trends, making it very difficult for editors and buyers to analyze them manually. On the other hand, the fast fashion companies such as ZARA and H&M need to change their product line faster. They have to generate accurate trends prediction with less time and labor costs in hope to reduce overstock. It is reported that H&M leaves $4.3 Billion in unsold inventory (Paton, 2018). The developed A.I. augmented fashion trends analysis methods showed that the A.I. model could analyze the fashion attributes and summarize the trends automatically without any labor costs involved.

The key findings of this exploratory study with the narrowly defined scope of work suggest future studies can be done to improve the accuracy in detecting fashion attributes in a wider range of fashion images in diverse settings.

**1)** *Street snapshots.* Only the runway shows and the online shopping photos were analyzed in this study. In future studies, street style images should also be analyzed using the proposed A.I. method. Street style is less dramatic than runway style. It may be interesting to compare the results from both street and runway images: for instance, street style images might be less diverse; there are no exaggerated decorations that could be observed from runway style images.

**2)** *Diversity.* In terms of analyzing fashion trends, this study only compared the results between Vogue and ZARA. More fashion brands and trend forecasting companies should be included in future studies. It would be interesting to observe the trend difference between the trend reports made by these fashion media/forecasting companies (Vogue and WGSN) and the products sold by these world's leading retailers (ZARA and H&M). It might also show whether the trend reports from Vogue and WGSN really affect the direction of products lines from these fashion brands.

**3)** *Unseen trends.* Due to the lack of location and geological information from the images, the proposed A.I. method was not able to explore unexpected or interesting trends that might exist in the images (as mentioned in 'Stage 2: Analyzing fashion trends'). Other than discovering some obvious fashion trends, the A.I. augmented method also has the potential to capture the complexity and multidimensional aspect of the image that cannot be observed visually.

    **4)** *Accuracy.* The proposed A.I. method was not always accurate. For example, Several inaccuracies were observed while analyzing the Dolce & Gabbana Spring 2018 ready-to-wear collection.

    In summary, research results show that the partnership with artificial intelligence algorithm gives "fashion insiders" a second pair of creative eyes. Certainly, it can lessen the financial and environmental burden of traditional labor-intensive abstraction process within fashion companies. A.I. algorithm has the potential to provide the fashion industry with a new tool improving the efficiency overall.

# 7. Supplementary Material

## 7.1 More results of fashion attributes detection on runway shows Spring/Summer 2018

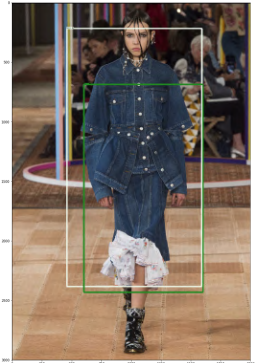

**Alexander Mcqueen**
detected textures:
 graphic: 60%
 print: 64%
detected fabrics:
 **denim: 98%**
detected shapes:
 **pencil: 50%**
 **bodycon: 75%**
detected parts:
 **button: 59%**
detected style:
 boyfriend: 54%
detected category: Dress: 93%
detected category: **Skirt: 50%**

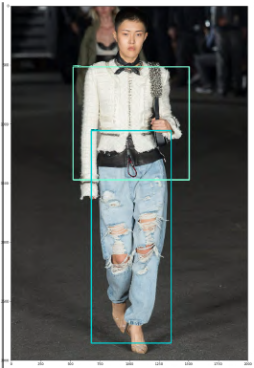

**Alexander Wang**
detected textures:
detected fabrics:
 chiffon: 60%
 **knit: 57%**
 crochet: 52%
 **denim: 60%**
 lace: 63%
 **fabric_beaded: 80%**
 **fabric_distressed: 82%**
detected shapes:
 high_rise: 53%
 **skinny: 65%**
 pullover: 56%
 ...

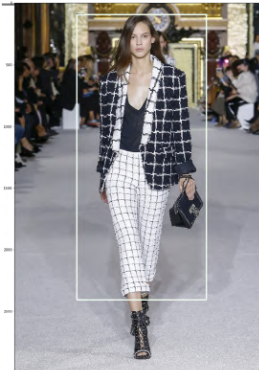

**Balmain**
detected textures:
 **tribal: 76%**
 **graphic: 55%**
 print: 77%
 dot: 66%
detected fabrics:
 **knit: 67%**
detected shapes:
 **crop: 54%**
 **midi: 52%**
detected parts:
 **sleeve: 50%**
detected style:
 rose: 59%
detected category: Dress: 84%

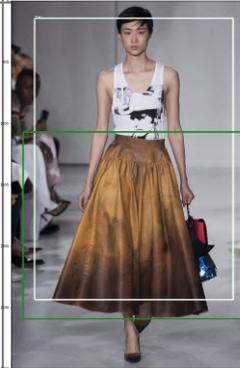

**Calvin Klein**
detected textures:
 **print: 56%**
detected fabrics:
 chiffon: 58%
 denim: 55%
 lace: 53%
detected shapes:
 **maxi: 62%**
 **muscle: 84%**
 pullover: 70%
 **crop: 66%**
 slim: 51%
detected parts:
 **sleeveless: 50%**
 ...

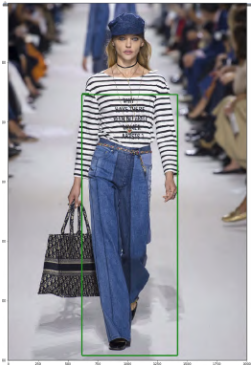

**Christian Dior**
detected textures:
 **stripe: 57%**
 graphic: 65%
 **stripe_nautical: 67%**
 pattern: 55%
 print: 63%
detected fabrics:
 **wash_acid: 71%**
 **denim: 74%**
 knit: 57%
 wash: 57%
detected shapes:
 crop: 57%
detected parts:
 cutout: 61% ...

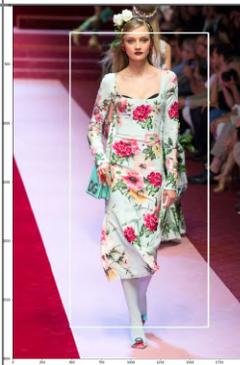

**Dolce & Gabbana**
detected textures:
 **floral_pattern: 80%**
 **graphic: 52%**
 **pattern_abstract: 56%**
 **print: 61%**
 **print_abstract: 82%**
detected fabrics:
 **chiffon: 72%**
 knit: 60%
detected shapes:
 **midi: 65%**
detected parts:
detected style:
 **summer: 60%** ...

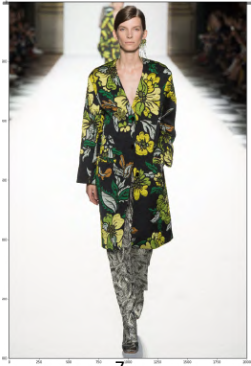

**Dries Van Noten**
detected textures:
 print_animal: 52%
 **floral_pattern: 62%**
 **tribal: 58%**
 **pattern: 50%**
 **print: 95%**
detected fabrics:
 fabric_beaded: 53%
 knit: 58%
detected shapes:
 **crop: 51%**
detected parts:
 edge_raw: 77%
 button: 61% ...

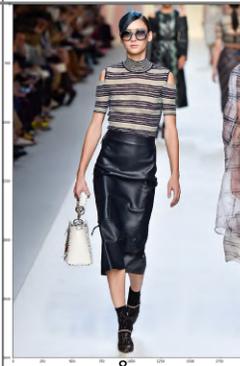

**Fendi**
detected textures:
 **graphic: 51%**
 floral_pattern: 59%
 **stripe: 52%**
 **print: 70%**
detected fabrics:
 **leather_faux: 67%**
 chiffon: 55%
 **knit_cable: 59%**
 denim: 54%
 lace: 90%
detected shapes:
 maxi: 62%
 **mid_rise: 52%**
 **midi: 66%** ...

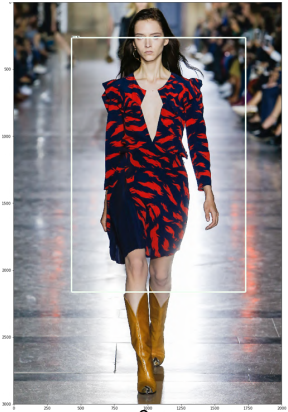

### Givenchy
<u>detected textures:</u>
**print: 84%**
**print_abstract: 71%**
dot: 58%
<u>detected fabrics:</u>
**cotton: 80%**
denim: 64%
**plaid: 76%**
<u>detected shapes:</u>
midi: 57%
<u>detected parts:</u>
sleeveless: 56%
<u>detected style:</u>
rose: 74%
**party: 73%** ...

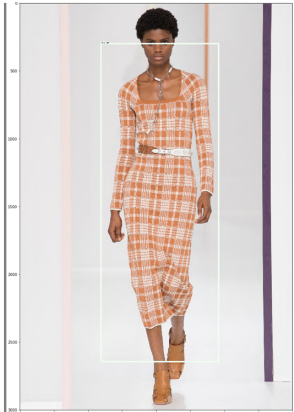

### Hermes
<u>detected textures:</u>
floral_pattern: 80%
**print: 62%**
<u>detected fabrics:</u>
denim: 56%
lace: 50%
<u>detected shapes:</u>
maxi: 55%
**midi: 51%**
**bodycon: 65%**
**pencil: 67%**
**crop: 72%**
<u>detected parts:</u>
patched: 52% ...

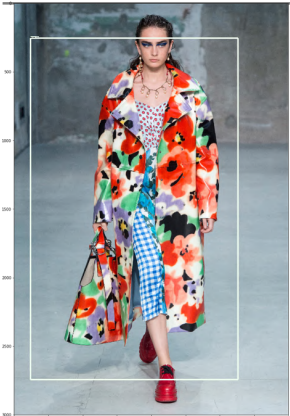

### Marni
<u>detected textures:</u>
**floral_pattern: 59%**
**pattern_abstract: 74%**
**print: 94%**
<u>detected fabrics:</u>
knit: 69%
<u>detected shapes:</u>
**midi: 53%**
pullover: 52%
<u>detected parts:</u>
<u>detected style:</u>
**rose: 53%**
**summer: 51%**
**tropical: 61%**
<u>detected category:</u>Dress: 95%

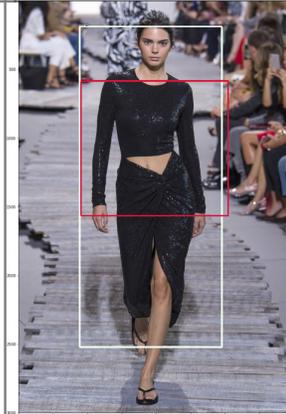

### Michael Kors
<u>detected textures:</u>
graphic: 72%
stripe: 94%
print: 90%
<u>detected fabrics:</u>
**knit: 78%**
<u>detected shapes:</u>
**crop: 59%**
**pencil: 75%**
**maxi: 63%**
**bodycon: 53%**
<u>detected parts:</u>
<u>detected style:</u>
rose: 55%
<u>detected category:</u>**Dress: 98%** ...

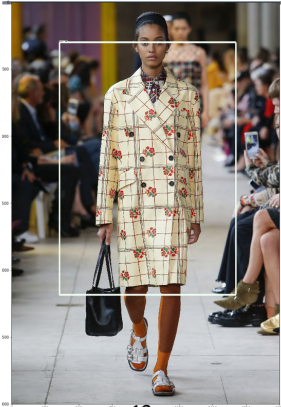

### Miu Miu
<u>detected textures:</u>
**graphic: 74%**
**print: 56%**
<u>detected fabrics:</u>
denim: 83%
**plaid: 72%**
<u>detected shapes:</u>
shift: 57%
<u>detected parts:</u>
edge_raw: 54%
<u>detected style:</u>
summer: 50%
<u>detected category:</u>Dress: 99%

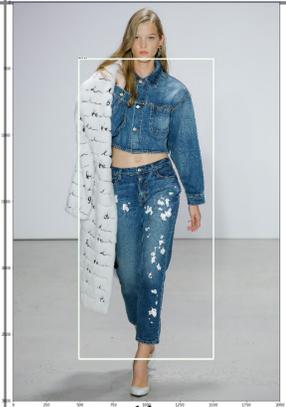

### Oscar de la Renta
<u>detected textures:</u>
paisley: 54%
graphic: 50%
dot: 52%
dot_polka: 65%
<u>detected fabrics:</u>
plaid: 50%
**denim: 53%**
**knit: 54%**
**fabric_distressed: 59%**
<u>detected shapes:</u>
**crop: 53%**
<u>detected parts:</u>
**sleeve: 68%** ...

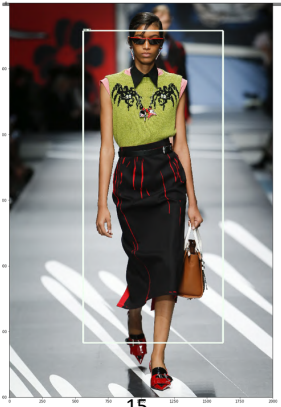

### Prada
<u>detected textures:</u>
floral_pattern: 85%
**graphic: 62%**
print: 58%
<u>detected fabrics:</u>
lace_floral: 64%
denim: 60%
lace: 66%
**knit: 52%**
<u>detected shapes:</u>
maxi: 55%
**midi: 59%**
**pencil: 53%**
pullover: 52%
**crop: 59%** ...

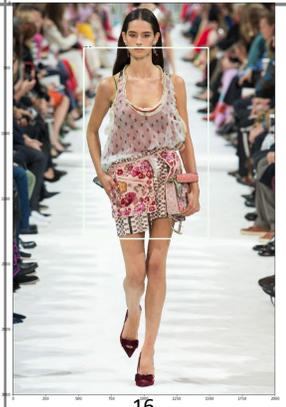

### Valentino
<u>detected textures:</u>
**floral_pattern: 62%**
**graphic: 56%**
**print: 59%**
<u>detected fabrics:</u>
denim: 56%
**knit: 56%**
<u>detected shapes:</u>
skater: 51%
**mini: 81%**
<u>detected parts:</u>
<u>detected style:</u>
**beach: 56%**
**summer: 61%**
<u>detected category:</u>Dress: 98%

**7.2 The descriptive statistics of the detected fashion attribute (for each category) on runway shows Spring/Summer 2018**

*7.2.1 The result of the detected textures on runway shows Spring/Summer 2018*

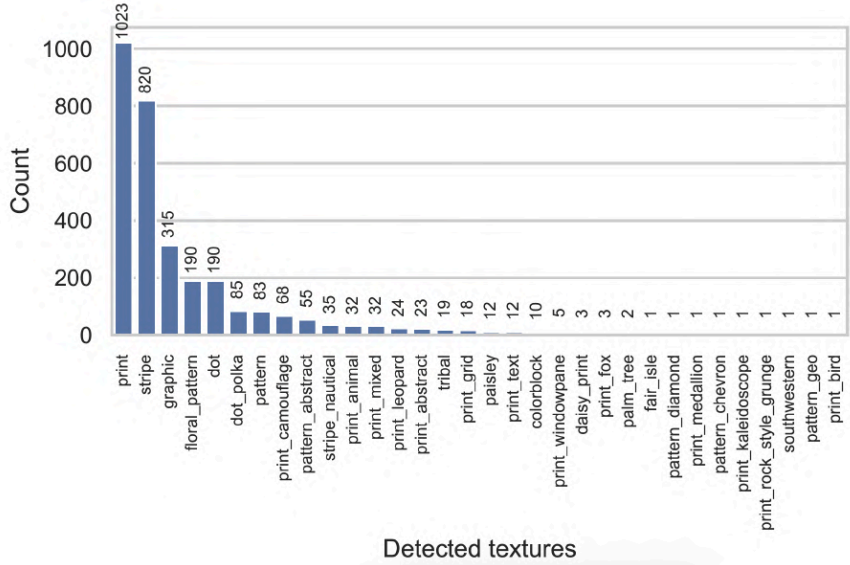

(Y axis indicates how many times the attributes have been detected by our A.I. model on runway shows Spring/Summer 2018)

*7.2.2 The result of the detected fabrics on runway shows Spring/Summer 2018*

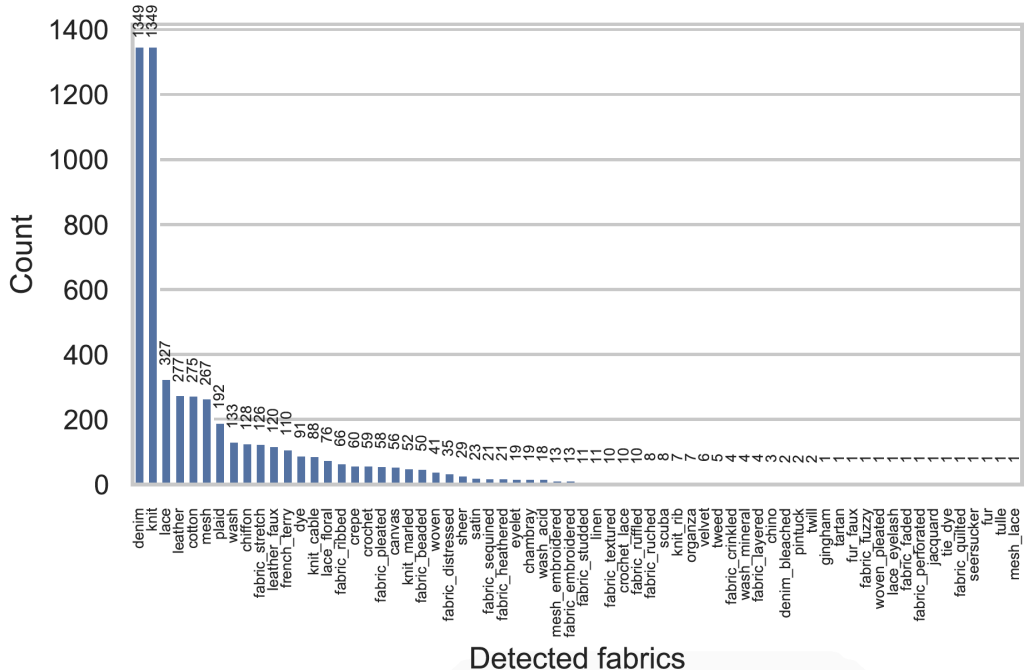

(Y axis indicates how many times the attributes have been detected by our A.I. model on runway shows Spring/Summer 2018)

*7.2.3 The result of the detected shapes on runway shows Spring/Summer 2018*

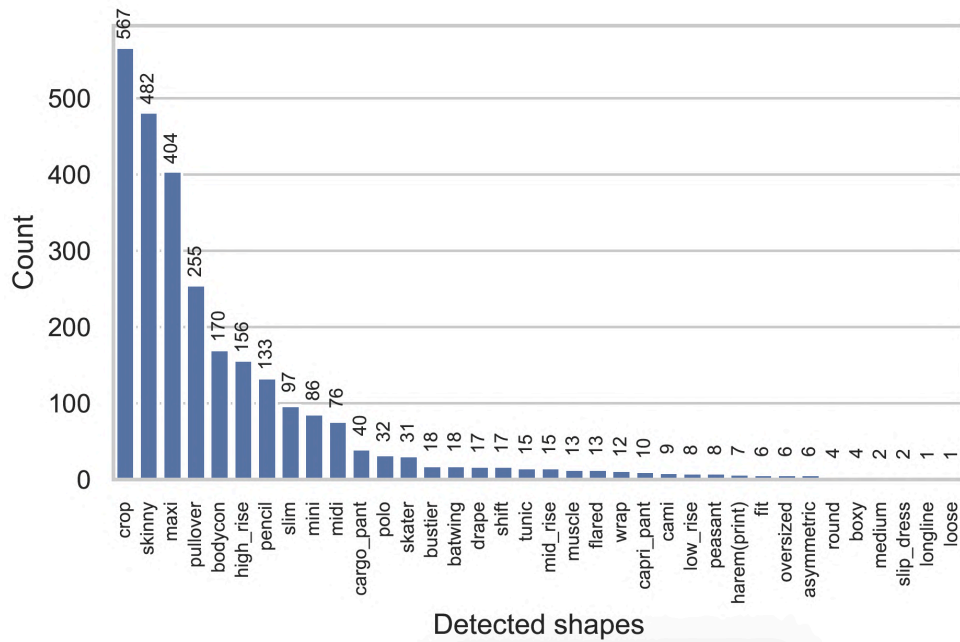

(Y axis indicates how many times the attributes have been detected by our A.I. model on runway shows Spring/Summer 2018)

*7.2.4 The result of the detected parts on runway shows Spring/Summer 2018*

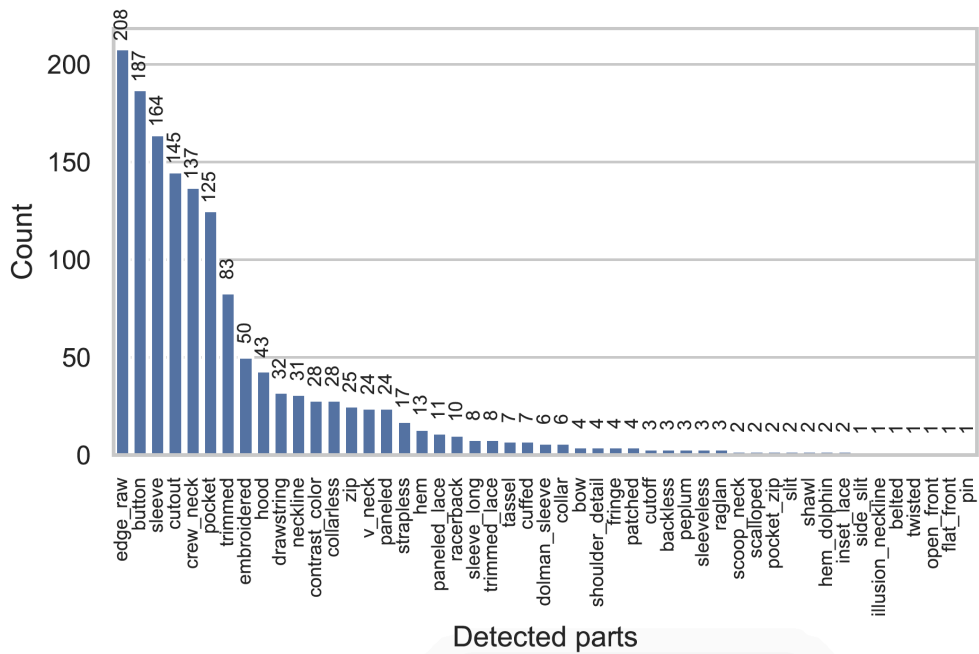

(Y axis indicates how many times the attributes have been detected by our A.I. model on runway shows Spring/Summer 2018)

*7.2.5 The result of the detected style on runway shows Spring/Summer 2018*

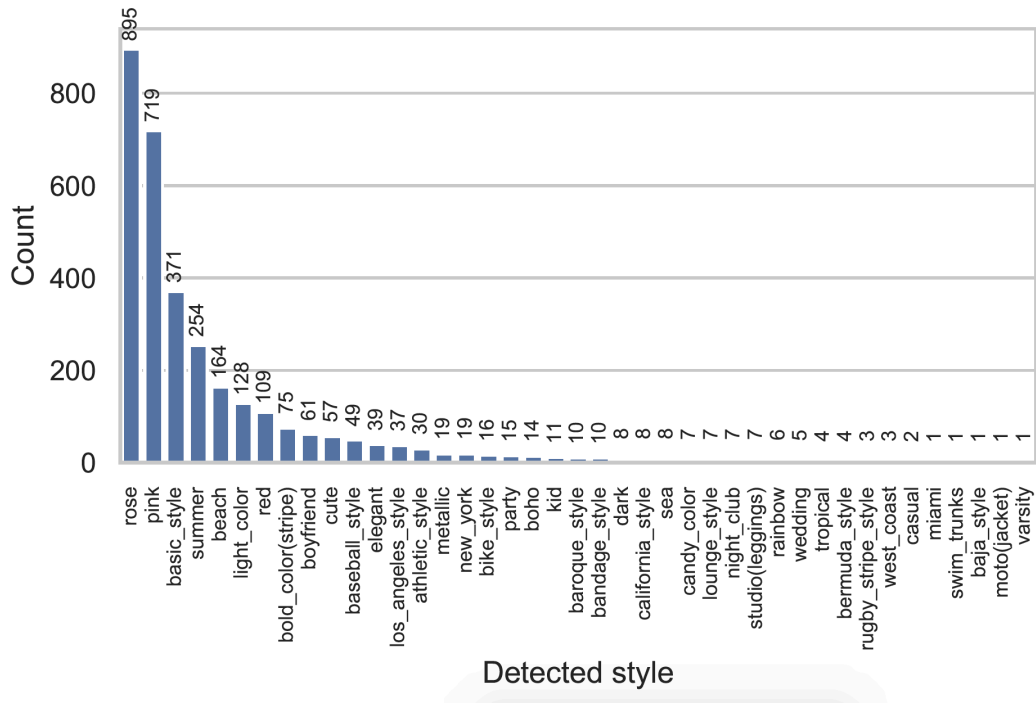

(Y axis indicates how many times the attributes have been detected by our A.I. model on runway shows Spring/Summer 2018)

*7.2.6 The result of the detected garment categories on runway shows Spring/Summer 2018*

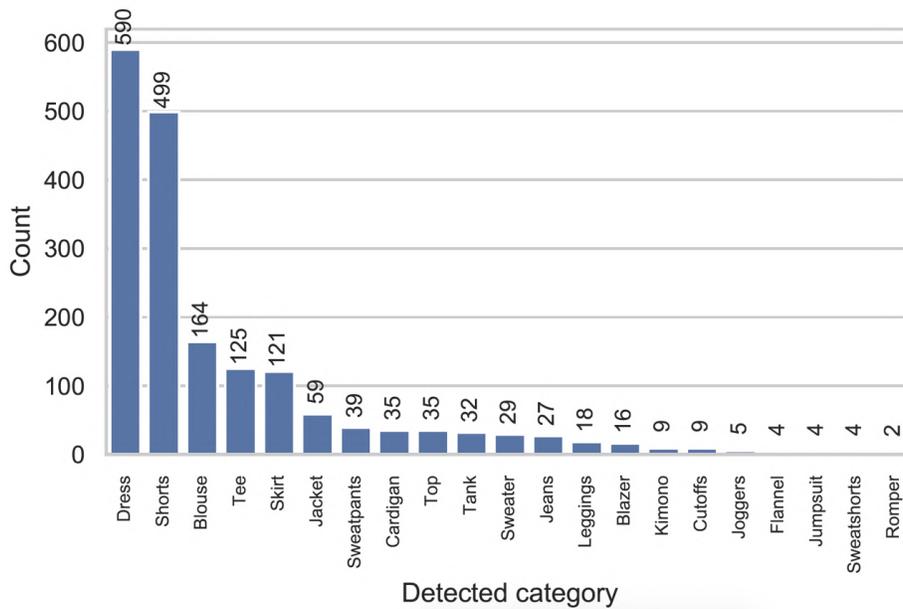

(Y axis indicates how many times the attributes have been detected by our A.I. model on runway shows Spring/Summer 2018)

## 7.3 A.I. model V.S. Vogue & ELLE & Harper's BAZAAR:

Compare the fashion trend prediction results from our A.I. model with the results from fashion editors on runway shows Spring/Summer 2018: In summary, the results show that the fashion editors missed a lot of attributes that our A.I. model has detected. A.I. model can analyze all of the runways images and provide a more comprehensive overview of all the fashion trends in a more efficient way, which we believe it's very difficult for any fashion editor to do so.

### 7.3.1 The mapping of attribute names among our A.I. model results, Vogue Italy, Vogue France, Vogue UK, ELLE, and Harper's BAZAAR.

| Mapped Name | Our A.I. Results | Vogue Italy | Vogue France | Vogue UK | ELLE | Harper's BAZAAR |
|---|---|---|---|---|---|---|
| Printed | Print | Head to Toe Prints | Printed | N/A | N/A | N/A |
| Floral | Floral pattern | N/A | Country garden (flower) | N/A | N/A | Coming Up Florals |
| Polka Dots | dot / dot polka | N/A | N/A | N/A | Black and White Polka Dots | N/A |
| Animal (pattern) | Print-animal | Animalier (Leopard, snakeskin, Tiger print) | N/A | N/A | N/A | N/A |
| Art Print | N/A | N/A | Arty impressions | N/A | N/A | Picasso Baby |
| Denim | Denim | N/A | Denim | Denim redux | Dark Denim Sets | Dark Denim |
| Check / Plaid (pattern) | Plaid | Check Patterns | Check | N/A | N/A | Mad for Plaid |
| Sheer | Sheer (chiffon) | Sheerness | N/A | Decent exposure | N/A | N/A |
| Glitter (fabric) | Fabric sequined | Glitter (sequins, crystals, lurex) | Glitter | N/A | N/A | N/A |
| Feathers | N/A | Feathers | Feathers | Extra texture (feathers) | Feathers | N/A |
| Plastic | N/A | N/A | Plastic | Plastic fantastic | Rubber / Transparent / Plastic | The Plastics |
| Pencil | Pencil | N/A | N/A | Pencil skirts | N/A | N/A |
| Asymmetric | N/A | Asymmetric | N/A | N/A | Asymmetrical Necklines | N/A |
| Trimming | Trimmed | Fringe | Fringes | Extra texture (fringing) | Fringe | Fringe Factor |
| Puffed / padded shoulder | N/A | N/A | 80s shoulders / shoulder pads | N/A | Puffed Shoulers | N/A |
| Pink | Pink | N/A | Pink | N/A | Pink and Red | Think Pink |
| Red | Red | N/A | Red | N/A | Pink and Red | N/A |
| Bold color | Bold color(stripe) | Splashes of colour | N/A | Crayola brights | N/A | N/A |
| Biker | Bike style | N/A | Biker | N/A | N/A | N/A |
| Lavender | N/A | N/A | N/A | N/A | Lavender | Paint The Town Lavender |

This table provides the mapped fashion attribute names among our A.I. model, Vogue, ELLE, and Harper's BAZZAR since the editors from these magazines might call them slightly different.

*7.3.2 The overlapped texture attributes among our A.I. model results, Vogue Italy, Vogue France, Vogue UK, ELLE, and Harper's BAZAAR.*

| Attribute Name | Our A.I. Results | Vogue Italy | Vogue France | Vogue UK | ELLE | Harper's BAZAAR |
|---|---|---|---|---|---|---|
| Printed | 1023 | ✓ | ✓ | | | |
| Floral | 190 | | ✓ | | | ✓ |
| Polka Dots | 85 | | | | ✓ | |
| Animal (pattern) | 32 | ✓ | | | | |
| Art Print | N/A | | ✓ | | | ✓ |
| Butterfly | N/A | | ✓ | | | |
| Rainbow | N/A | | | | ✓ | |

Across 1844 runway images across 46 collections of the Spring/Summer 2018 presented in New York, London, Milan and Paris, our A.I. model finds attribute 'printed' among 1023 runway images and indicates it is the most popular attributes in 'texture' category. One the editorial sides, Vogue Italy and Vogue France also indicate 'printed' as the popular attributes for Spring/Summer 2018. However, Vogue UK, ELLE, and Harper's BAZAAR didn't mention anything about 'printed' in their magazines.

*7.3.3 The overlapped fabric attributes among our A.I. model results, Vogue Italy, Vogue France, Vogue UK, ELLE, and Harper's BAZAAR.*

| Attribute Name | Our A.I. Results | Vogue Italy | Vogue France | Vogue UK | ELLE | Harper's BAZAAR |
|---|---|---|---|---|---|---|
| Denim | 1349 | | ✓ | ✓ | ✓ | ✓ |
| Check / Plaid (pattern) | 192 | ✓ | ✓ | | | ✓ |
| Sheer | 29 | ✓ | | ✓ | | |
| Glitter (fabric) | 21 | ✓ | ✓ | | | |
| Feathers | N/A | ✓ | ✓ | ✓ | ✓ | |
| Plastic | N/A | | ✓ | ✓ | ✓ | ✓ |

Except Vogue Italy, Our A.I. model, Vogue France, Vogue UK, ELLE, and Harper's BAZAAR indicate 'demin' is the most popular fabric attributes for the Spring/Summer 2018.

*7.3.4 The overlapped shape attributes among our A.I. model results, Vogue Italy, Vogue France, Vogue UK, ELLE, and Harper's BAZAAR.*

| Attribute Name | Our A.I. Results | Vogue Italy | Vogue France | Vogue UK | ELLE | Harper's BAZAAR |
|---|---|---|---|---|---|---|
| Pencil | 133 | | | ✓ | | |
| Asymmetric | 6 | ✓ | | | ✓ | |

Only Our A.I. model and Vogue UK suggest that 'pencil' is the most popular shape attribute attributes for the Spring/Summer 2018. Our A.I. model find 'pencil' from 133 runway images. The editors from Vogue Italy, Vogue France, ELLE, and Harper's BAZAAR missed this important attribute.

*7.3.5 The overlapped parts attributes among our A.I. model results, Vogue Italy, Vogue France, Vogue UK, ELLE, and Harper's BAZAAR.*

| Attribute Name | Our A.I. Results | Vogue Italy | Vogue France | Vogue UK | ELLE | Harper's BAZAAR |
|---|---|---|---|---|---|---|
| Trimming | 83 | ✓ | ✓ | ✓ | ✓ | ✓ |
| Puffed / padded shoulder | N/A | | ✓ | | ✓ | |
| Veils | N/A | | ✓ | | | |
| Ruching | N/A | | | | ✓ | |
| Square Necklines | N/A | | | | ✓ | |
| Embellished Straps | N/A | | | | ✓ | |

Both our A.I. model and all magazines agree that 'trimming' is the popular parts attributes for the Spring/Summer 2018. However, the editors also suggest the other attributes such as 'puffed / padded shoulder' are the popular trends, which are not detected by our A.I. model.

*7.3.1 The overlapped style attributes among our A.I. model results, Vogue Italy, Vogue France, Vogue UK, ELLE, and Harper's BAZAAR.*

| Attribute Name | Our A.I. Results | Vogue Italy | Vogue France | Vogue UK | ELLE | Harper's BAZAAR |
|---|---|---|---|---|---|---|
| Pink | 719 | | ✓ | | ✓ | ✓ |
| Red | 109 | | ✓ | | ✓ | |
| Bold color | 75 | ✓ | | ✓ | | |
| Biker | 16 | | ✓ | | | |
| Lavender | N/A | | | | ✓ | ✓ |
| Cowgirl | N/A | | ✓ | | | |
| Tropical rhythm | N/A | | ✓ | | | |
| Night Fever | N/A | | ✓ | | | |
| Ice cream girl | N/A | | | ✓ | | |
| Haute Performance | N/A | | | | ✓ | |
| The Virgin Suicides | N/A | | | | | ✓ |
| Work It Out | N/A | | | | | ✓ |

It seems not only there is a big disagreement among our A.I. models and all the magazines, but also the editors from different magazines have very different opinions on what is the most popular color for the Spring/Summer 2018.

## Acknowledgements

We thank Serge Belongie, Claire Cardie, Kavita Bala, Wei Cao, and Menglin Jia for their helpful feedback and discussion in the development of this paper. We also thank R.J. Lehman, Corina and Dan Lecca for providing the runway show images for this study.